\documentclass[10pt,twocolumn,letterpaper]{article}

\usepackage{cvpr}              
\definecolor{cvprblue}{rgb}{0.21,0.49,0.74}
\usepackage[pagebackref,breaklinks,colorlinks,allcolors=cvprblue]{hyperref}
\usepackage{multirow}
\usepackage{caption}
\newcommand{\std}[1]{{\scriptsize$\pm$#1}}

\def\confName{CVPR}
\def\confYear{2027}

\title{WM-Quant: Rollout-Aware Post-Training Quantization for JEPA-Style World Models}

\author{
William Hammer\thanks{Equal contribution.}, \quad Andrew Heuer\footnotemark[1], \quad  Nikhil Krishna,  \quad Matthew Grenier, \quad Ramtin Zand\\
University of South Carolina\\
Columbia, SC, USA\\
{\tt\small WHAMMER@email.sc.edu, AHEUER@email.sc.edu, NKRISHNA@email.sc.edu} \\
{\tt\small  MGRENIER@email.sc.edu, RAMTIN@cse.sc.edu}}

\begin{document}
\maketitle
\begin{abstract}
We present WM-Quant, to our knowledge, the first post-training quantization (PTQ) framework designed specifically for JEPA-style world models. Although world models typically feature relatively few parameters, action planning algorithms often require thousands of inferences, which can quickly become computationally expensive. PTQ is a natural way to reduce this cost. However, we show that in world models using autoregressive model predictive control (MPC) rollouts, quantization error can compound with each predictor step, resulting in action rankings from the quantized model becoming near-random. WM-Quant addresses this with three components that build on existing PTQ methods: early-encoder precision, control-path preservation, and rollout-aware scaling. Early-encoder precision protects accurate latent state and goal information to mitigate long-term error propagation. Control-path preservation allows the world model predictor to distinguish between fine-grained actions, significantly improving action rankings and execution. Rollout-aware scaling jointly calibrates encoder and per-step predictor activation ranges with respect to an entire rollout rather than a single prediction, reducing the cumulative rollout error. Across multiple world models, environments, and PTQ methods, WM-Quant substantially improves planning success at low bit-widths, approaching full-precision performance at W4A4.

\end{abstract}
 
\begin{figure}[t]
  \centering
  \includegraphics[width=\columnwidth]{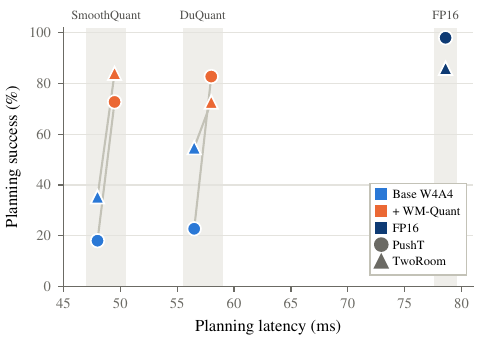}
  \caption{Planning latency vs.\ planning success on LeWM (W4A4), averaged over 3 seeds. WM-Quant approaches floating point results on both TwoRoom and PushT datasets with little impact on latency.}
  \label{fig:latency_success}
\end{figure}

\section{Introduction}

World models \cite{lecun2022path} allow an embodied agent to predict how their environment will respond to chosen actions \cite{zidan2026worldmodelscomprehensivesurvey}. Learned dynamics and an internal simulator allow for both forward and counterfactual rollouts, essential for next-state prediction and decision-making \cite{zidan2026worldmodelscomprehensivesurvey}. Recent advancements in Joint Embedding Predictive Architectures (JEPA) \cite{lecun2022path}, where world models reason in latent space, such as in LeWorldModel (LeWM) \cite{maes2026leworldmodelstableendtoendjointembedding}, DINO-WM \cite{zhou2025dinowm}, or V-JEPA 2-AC \cite{assran2025vjepa2selfsupervisedvideo}, have allowed world models to discard task-irrelevant detail \cite{lecun2022path} and reason faster than pixel-space alternatives \cite{assran2025vjepa2selfsupervisedvideo}. 

Despite these gains, JEPA-style world models still may not be adequate for time-critical domains such as robotics and autonomous driving, where predicting future states in a timely manner is crucial. This is because world models must evaluate many candidate action sequences at each decision step, making inference latency a key bottleneck for real-time planning deployment \cite{fu2026empiricalstudyworldmodel}. On PushT \cite{10.1177/02783649241273668, florence2021implicit}, for example, a dense latent world model requires roughly two minutes per planning decision on a single GPU \cite{yin2026ddpwm}. In addition, many world model applications are targeted towards devices that are highly memory-constrained \cite{gholami2021surveyquantizationmethodsefficient}. Recent efforts to address these issues fall into several directions, such as quantization \cite{fu2026empiricalstudyworldmodel}, sparse computation \cite{yin2026ddpwm, ICLR2026_a750d522}, and compact architectures \cite{maes2026leworldmodelstableendtoendjointembedding, nam2026causaljepa}. 

Recent work applies standard post-training quantization (PTQ) methods to DINO-WM \cite{zhou2025dinowm} and finds that low-bit quantization can substantially degrade planning performance, with the encoder particularly sensitive to quantization \cite{fu2026empiricalstudyworldmodel}. However, it does not propose a method to address these limitations. Additionally, the evaluated methods are loop-invariant, and do not account for error propagation across rollout steps. Recent work on world action models \cite{zhou2026quantwamscalibratingrightgranularity} targets diffusion-based video–action generation, where error accumulates over denoising steps rather than through latent planning rollouts. To our knowledge, no quantization method has been designed specifically to limit error propagation in latent world model planning.

In this work, we propose WM-Quant, a unified static PTQ framework that can be applied on top of existing static PTQ methods to substantially improve the planning performance of world models. We first define the error propagation problem with respect to world model rollouts, rather than focusing solely on single prediction performance. Guided by this analysis, WM-Quant selects early encoder blocks and control information to retain at higher bit-widths, reducing error propagation. Per-rollout-step predictor scales and encoder activation scaling factors are learned jointly. When evaluated on LeWM PushT, WM-Quant achieves a 58\% success rate at W3A3 and an 81.3\% success rate at W4A4, compared to baselines of just 10\% and 20\%. The main contributions of the paper are:

\begin{enumerate}
    \item A detailed analysis of error propagation in quantized planning-oriented world models.
    \item WM-Quant, to our knowledge, the first PTQ framework designed \textit{specifically} for JEPA-style world models, achieving state-of-the-art performance under extreme low-precision inference.
\end{enumerate}

\section{Related work}
\subsection{World models}
The line of world model work we build upon originates from \cite{lecun2022path}, which proposes a broad strategy for training autonomous agents centered on a world model that predicts an action's next state. It goes on to propose JEPA for world models, where a predictor can make abstract predictions on a compressed world state in latent space, rather than being required to predict pixel-level information. 

JEPA has been applied to representation learning tasks \cite{Assran_2023_CVPR, bardes2024revisitingfeaturepredictionlearning}, and action-conditioned world models for planning \cite{zhou2025dinowm, maes2026leworldmodelstableendtoendjointembedding, sobal2025learning}. These models typically plan with model predictive control (MPC) \cite{GARCIA1989335, MAYNE2000789}, using the cross-entropy method (CEM) \cite{10.1023/A:1010091220143} to optimize action sequences against a latent goal \cite{pmlr-v97-hafner19a, 3327345.3327385}. DINO-WM \cite{zhou2025dinowm} builds its latent space on a frozen, pretrained DINOv2 \cite{oquab2024dinov2learningrobustvisual} encoder, while LeWM \cite{maes2026leworldmodelstableendtoendjointembedding} trains the encoder and predictor end-to-end with a Gaussian regularizer to prevent representation collapse. On offline, goal-conditioned planning benchmarks, these approaches outperform reconstruction- or temporal-difference-based world models such as the Dreamer \cite{hafner2023mastering, hafner2021mastering, Hafner2020Dream} and TD-MPC \cite{hansen2022temporal, hansen2024tdmpc} models. However, MPC requires thousands of predictor evaluations per decision \cite{zhou2025dinowm, maes2026leworldmodelstableendtoendjointembedding, pmlr-v97-hafner19a}, making inference cost a key barrier to deployment. 

\subsection{Quantization}
To reduce inference costs, previous research has explored many different PTQ methodologies \cite{pmlr-v119-nagel20a, li2021brecq, wei2022qdrop, NEURIPS2022_1caf09c9, ashkboos2024quarot, liu2025spinquant, Mohammadi_2026_CVPR}. 
GPTQ \cite{frantar2023optq} uses calibration data to estimate second-order information and quantizes transformer weights one at a time, updating the remaining weights to compensate for quantization error. SmoothQuant \cite{pmlr-v202-xiao23c} introduces a per-channel, invertible, linear transformation which can be baked into model weights and distributes quantization error between weights and activations. OmniQuant \cite{shao2024omniquant} extends SmoothQuant by learning weight clipping and smoothing transformations with gradient descent. DuQuant \cite{lin2024duquant} introduces rotation matrices and zigzag permutations which spread extreme outliers across channels. Mixed-precision methods \cite{Wang_2019_CVPR} keep sensitive layers at higher bit-widths. Quantization of iterative models with continuous outputs has been studied for state space models \cite{chiang2024quambaposttrainingquantizationrecipe, chiang2025quamba2robustscalableposttraining}, looped language models \cite{fang2026loopqquantizationloopedlanguage}, world action models \cite{zhou2026quantwamscalibratingrightgranularity}, and most commonly diffusion \cite{shang2023posttrainingquantizationdiffusionmodels, so2023temporal, Li_2023_ICCV, he2023ptqdaccurateposttrainingquantization}, where error accumulates over the denoising trajectory. 
\begin{figure*}[t]
  \centering
  \includegraphics[width=0.85\textwidth]{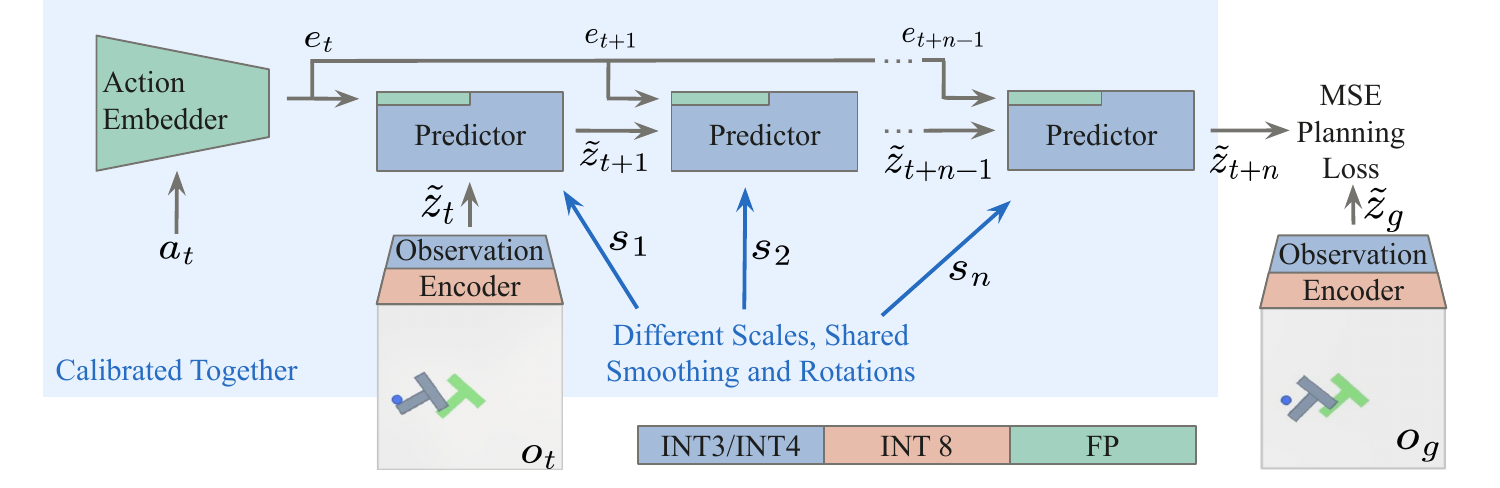}
  \caption{Overview of our proposed world model quantization framework. We preserve action information to improve candidate action ranking and the first three encoder blocks to limit early encoder error in the predictor. We also calibrate per-iteration scales with respect to an entire rollout, rather than a single prediction. These additions work together to limit error propagation and increase the rate at which models reach the goal state.}
  \label{fig:overview}
\end{figure*}

\section{Methodology}
\label{sec:methodology}

\subsection{Quantization preliminaries}
Quantization is the process of mapping continuous high-precision values to a fixed discrete set of numbers \cite{gholami2021surveyquantizationmethodsefficient}. This process is used to decrease the latency and memory usage of machine learning models by storing weights and activations as low-bit integers. To create a consistent testing configuration, we perform uniform symmetric per-channel quantization on weights and uniform asymmetric static per-tensor quantization on activations across all evaluated methods. Only asymmetric equations are shown for brevity. During quantization to a bit width $b$, each tensor is calibrated with a clipping range $[\alpha, \beta]$, from which a scale $s$ and zero-point $z$ are derived:
\begin{equation}
  s = \frac{\beta - \alpha}{2^b - 1}, \qquad
  z = \left\lfloor -\frac{\alpha}{s} \right\rceil .
\end{equation}
A real value $r$ is then quantized to an integer $q$ and dequantized to $\tilde{r}$ as
\begin{equation}
  q = \mathrm{clamp}\!\left( \left\lfloor \frac{r}{s} \right\rceil + z,\; 0,\; 2^b - 1 \right),
  \qquad
  \tilde{r} = s\,(q - z),
  \label{eq:quant}
\end{equation}
where $\lfloor \cdot \rceil$ is a rounding operation mapping a real-valued number to an integer.
The quantization error, $\tilde{r} - r$, has two sources: rounding error when $r$ is inside the range and clipping error when $r$ lies outside it.
$\alpha$ and $\beta$ are calibrated to minimize both sources of error.
Recent quantization methods also introduce smoothing transformations \cite{pmlr-v202-xiao23c, shao2024omniquant} and rotations \cite{ashkboos2024quarot, lin2024duquant, liu2025spinquant} which are often fused into neighboring layers and act to distribute outliers between weights and activations, and across channels.

\subsection{World model preliminaries}
In general, JEPA-style world models have three core components \cite{lecun2022path, zhou2025dinowm, maes2026leworldmodelstableendtoendjointembedding}. The first is an encoder, which takes an observation $o_t$ at timestep $t$ and compresses it into a semantically rich latent state representation $z_t$. The second is an action embedder, which takes an agent action $a_t$ and transforms it into a high-dimensional vector representation $e_t$. The third and final component is the predictor, which predicts the latent state at timestep $t+1$ from the previous $H$ state representations and action embeddings. This can be represented formally as:
\begin{equation}
\begin{aligned}
    \text{Observation encoder:} &\quad z_t = \mathrm{enc}_\theta(o_t), \\
    \text{Action embedder:}     &\quad e_t = \mathrm{emb}_\phi(a_t), \\
    \text{Predictor:}           &\quad \hat{z}_{t+1} = \mathrm{pred}_\psi\big(\hat{z}_{t-H+1:t},\, e_{t-H+1:t}\big),
\end{aligned}
\end{equation}
where $\hat{z}_t$ is used to represent a predicted state. When deployed, these models run autoregressively for $n$ steps, which is referred to as a rollout. A rollout begins at some timestep $\tau$ (e.g.,\ the current timestep when replanning): observations up to $\tau$ are encoded, $\hat{z}_t = z_t$ for $t \le \tau$, and the predictor is then applied to its own outputs for $t \ge \tau$. Information from these rollouts can be used to guide agent actions or can be decoded to predict future observations.

A common approach to planning with world models is MPC \cite{GARCIA1989335, MAYNE2000789}, which uses observations of the current environment state, a predictive model, and an optimization method to find an action sequence that minimizes a cost function. It then executes one or more optimized actions and repeats the process. The optimization method we leverage is the cross-entropy method (CEM) \cite{10.1023/A:1010091220143}, in which the planner samples $c$ candidate action sequences from a Gaussian distribution. The top-$k$ sampled action sequences with the lowest cost are then selected, and the distribution is updated based on the selected action sequences. This process repeats for $m$ iterations, after which some number of actions from the final mean action sequence are executed.

In many world model benchmarks, including those we test on \cite{10.1177/02783649241273668, florence2021implicit, sobal2025learning, zhou2025dinowm}, some goal state $g$ is used as an objective and the difference between the current and goal latent states is used as the cost function \cite{zhou2025dinowm, maes2026leworldmodelstableendtoendjointembedding}. Similar to CEM, this is highly convenient for model evaluation, but is not necessary to implement our framework.

\begin{figure}[t]
  \centering
  \includegraphics[width=\linewidth, clip]{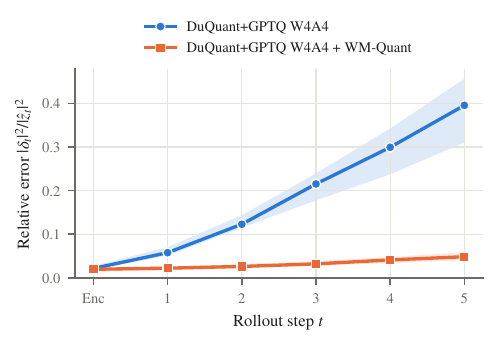}
  \caption{Latent error over a planning rollout on LeWM PushT (W4A4), relative to the full-precision model. Step 0 is the encoded initial state; steps 1–5 are successive predictor outputs. Lines show the median over start states and shaded regions the interquartile range. The rate of error propagation with WM-Quant is substantially slower than without it.} 
  \label{fig:growth}
\end{figure}

\subsection{Error propagation}
\label{sec:error_prop}
In our view, the key issue of world model quantization is error propagation. During a rollout, the predictor is conditioned on its own previous outputs, so when the model is quantized, each predictor step receives inputs that already contain error and introduces its own error (Fig. \ref{fig:growth}). This effect can be modeled by an error term added to the output of each component, where $\eta_t$, $\zeta_t$, and $\epsilon_t$ denote the error introduced by quantizing the encoder, action embedder, and predictor, respectively. The quantized rollout is then
\begin{equation}
\begin{aligned}
    \tilde{z}_t     &= \mathrm{enc}_\theta(o_t) + \eta_t, && t \le \tau \\
    \tilde{e}_t     &= \mathrm{emb}_\phi(a_t) + \zeta_t \\
    \tilde{z}_{t+1} &= \mathrm{pred}_\psi\big(\tilde{z}_{t-H+1:t},\, \tilde{e}_{t-H+1:t}\big) + \epsilon_{t+1}, && t \ge \tau
\end{aligned}
\end{equation}
Let $\delta_t = \tilde{z}_t - \hat{z}_t$ denote the accumulated error at step $t$, so that $\delta_t = \eta_t$ for $t \le \tau$. Omitting the window subscripts $t-H+1{:}t$, the error at each subsequent step is
\begin{equation}
    \delta_{t+1} = \underbrace{\mathrm{pred}_\psi(\tilde{z}, \tilde{e}) - \mathrm{pred}_\psi(\hat{z}, e)}_{\text{propagated error}}
                 + \underbrace{\epsilon_{t+1}}_{\text{ predictor error}}
\end{equation}
To model how error propagates, we linearize the predictor around the full-precision rollout. This is an accurate approximation when errors are small, but acts as a heuristic in cases with large errors. For clarity, we present the analysis for a predictor with $H=1$.
Let
\begin{equation}
  A_t = \frac{\partial\,\mathrm{pred}_\psi}{\partial z}\Big|_{(\hat z_t,\, e_t)},
  \qquad
  B_t = \frac{\partial\,\mathrm{pred}_\psi}{\partial e}\Big|_{(\hat z_t,\, e_t)}
\end{equation}
denote the Jacobians of the predictor with respect to its state and action inputs.
To first order, the error then evolves as
\begin{equation}
  \delta_{t+1} \approx
  \underbrace{A_t\,\delta_t}_{\text{propagated error}}
  \;+\;
  \underbrace{B_t\,\zeta_t + \epsilon_{t+1}}_{\text{new error}},
  \qquad \delta_\tau = \eta_\tau .
  \label{eq:rec}
\end{equation}
Each step therefore transforms the error inherited from the previous step through the predictor’s state Jacobian and adds the error introduced by quantizing the action embedder and predictor at that step.

\paragraph{Unrolling the rollout.}
Substituting Eq.~\eqref{eq:rec} into itself,
\begin{align*}
  \delta_{\tau+1} &\approx A_\tau\,\eta_\tau + \big(B_\tau\zeta_\tau + \epsilon_{\tau+1}\big), \\
  \delta_{\tau+2} &\approx A_{\tau+1}A_\tau\,\eta_\tau
                    + A_{\tau+1}\big(B_\tau\zeta_\tau + \epsilon_{\tau+1}\big) \\
                    &+ \big(B_{\tau+1}\zeta_{\tau+1} + \epsilon_{\tau+2}\big),
\end{align*}
so every error is multiplied by the state Jacobian of each step that follows it.
Writing $\Phi_{t:s} = A_{t-1}A_{t-2}\cdots A_s$ for the propagator from step $s$
to step $t$ (with $\Phi_{t:t} = I$), the error after an $n$-step rollout is
\begin{equation}
  \delta_{\tau+n} \approx
  \underbrace{\Phi_{\tau+n:\tau}\,\eta_\tau}_{\text{encoder}}
  \;+\;
  \sum_{s=\tau}^{\tau+n-1} \Phi_{\tau+n:s+1}
  \big(\underbrace{B_s\,\zeta_s}_{\text{embedder}}
  + \underbrace{\epsilon_{s+1}}_{\text{predictor}}\big).
  \label{eq:unrolled}
\end{equation}

It is worth noting the distinction between the compounding error problem in world models and large language models (LLMs). While both model types are autoregressive, LLMs discretize their outputs into a set of tokens and continuous activations never pass through the model more than once. Thus, it is possible to add a significant amount of error in an LLM inference step and still receive an output token identical to the unperturbed one \cite{dutta2024accuracyneed}, effectively resetting $\delta_t$ to zero. Because JEPA world models condition directly on information-dense latent embeddings, no such reset occurs, and error propagation poses a central challenge (Fig.~\ref{fig:growth}).

\subsection{WM-Quant framework}
\label{sec:WM-Quant}
To address the ways error propagates throughout a rollout, we introduce WM-Quant, a world model quantization framework consisting of three key components, each of which tackles a different source of error (Fig. {\ref{fig:overview}). Taken together, WM-Quant provides a blueprint that can be applied to any quantization method for JEPA-style world models.

\begin{figure*}[t]
  \centering
  \includegraphics[width=\textwidth]{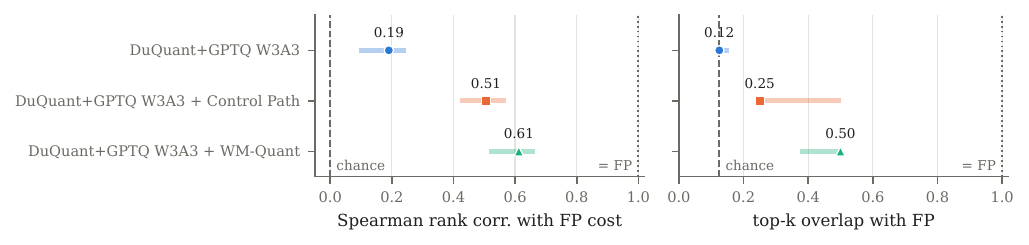}
  \caption{Agreement between the quantized and full-precision rankings of 64 CEM candidate
           action sequences on LeWM PushT (W3A3), median over 16 start states; bars show the
           interquartile range. Dashed line: chance level; dotted line: identical to FP. Top-8 overlap is displayed.}
  \label{fig:ranking}
\end{figure*}

\paragraph{Early-encoder precision}

Quantization error in the encoder is important for two reasons. The first is that it affects the model twice: once, in the hidden state $\hat{z}_t$ and once in the goal state, which decides which states are selected in CEM optimization. The second is that the encoder error term is propagated throughout the entire rollout, as expressed in Eq. (\ref{eq:unrolled}). This means even if $\eta_t$ is a small perturbation in the encoder, it contains an outsized potential to introduce large compounded error throughout the rollout. Within the encoder, the earliest blocks tend to be the most sensitive to quantization \cite{ranjan2024lrpqvitmixedprecisionvisiontransformer}. WM-Quant therefore keeps the activations of the first three encoder blocks at 8 bits in all experiments, while their weights remain at the target bit-width. All remaining blocks use the target bit-width for both weights and activations. This change is inexpensive because the encoder processes only observed frames and the goal, whereas the predictor is invoked at every rollout step for every candidate action sequence. On Wall \cite{zhou2025dinowm}, for example, each replanning step encodes two frames, while the predictor is evaluated 15,000 times: 300 candidates × 10 CEM iterations × a 5-step horizon. In some deployment scenarios, it may even be optimal to keep the encoder at full precision, though this would have a large memory impact.

\paragraph{Control-path preservation}
Quantization error in the action-conditioning path is particularly detrimental to world model planning. As part of CEM, many action sequences are sampled from a Gaussian distribution, resulting in closely related actions. The differences in these actions are what separates each candidate rollout, highlighting the importance of embedding fidelity. Even a relatively small $\zeta_t$ can perturb the action embeddings dramatically enough to change the ranking order and shift the action distribution further from the intended goal (Fig.~\ref{fig:ranking}). This is compounded by the fact that, as shown in Eq.~\eqref{eq:unrolled}, $\zeta_t$ is introduced at each rollout step. 

Furthermore, in some models such as DINO-WM, low-magnitude action and proprioceptive features are attached directly to much larger latent state embeddings. This results in control information collapsing in the quantization process due to scale disparity. To account for these effects, WM-Quant keeps the control path in floating point, preserving the fine-grained action information necessary for planning. This reduces $\zeta_t$ to zero while affecting a small percentage of predictor parameters. On DINO-WM, this is $\sim1.5\%$ of model parameters, and on the much smaller LeWM it is $\sim7.5\%$.

\paragraph{Rollout-aware scaling}

\begin{table*}[t]
  \centering
  \footnotesize

  \caption{Planning success (\%) on LeWM PushT after the first (iter 1) and second (iter 2) planning
           iteration, mean over 3 seeds. $\Delta$ is +WM-Quant minus the base method.}
  \label{tab:pusht_mean}
  \begin{tabular*}{\textwidth}{@{\extracolsep{\fill}} l cccccc cccccc}
    \toprule
    & \multicolumn{6}{c}{W3A3} & \multicolumn{6}{c}{W4A4} \\
    \cmidrule(lr){2-7} \cmidrule(lr){8-13}
    & \multicolumn{2}{c}{Base} & \multicolumn{2}{c}{+WM-Quant} & \multicolumn{2}{c}{$\Delta$}
      & \multicolumn{2}{c}{Base} & \multicolumn{2}{c}{+WM-Quant} & \multicolumn{2}{c}{$\Delta$} \\
    \cmidrule(lr){2-3} \cmidrule(lr){4-5} \cmidrule(lr){6-7}
    \cmidrule(lr){8-9} \cmidrule(lr){10-11} \cmidrule(lr){12-13}
    Method & iter 1 & iter 2 & iter 1 & iter 2 & iter 1 & iter 2 & iter 1 & iter 2 & iter 1 & iter 2 & iter 1 & iter 2 \\
    \midrule
    FP           & 80.0 & 98.0 & -- & -- & -- & -- & 80.0 & 98.0 & -- & -- & -- & -- \\
    \midrule
    SmoothQuant  & 4.0  & 4.7  & \textbf{28.7} & \textbf{39.3} & +24.7 & +34.7
                 & 12.0 & 18.0 & \textbf{52.7} & \textbf{72.7} & +40.7 & +54.7 \\
    DuQuant      & 9.3  & 13.3 & \textbf{31.3} & \textbf{54.0} & +22.0 & +40.7
                 & 15.3 & 22.7 & \textbf{62.7} & \textbf{82.7} & +47.3 & +60.0 \\
    DuQuant+GPTQ & 8.7  & 10.0 & \textbf{41.3} & \textbf{58.0} & +32.6 & +48.0
                 & 12.7 & 20.0 & \textbf{66.0} & \textbf{81.3} & +53.3 & +61.3 \\
    RTN          & 2.0  & 5.3  & \textbf{27.3} & \textbf{38.0} & +25.3 & +32.7
                 & 8.7  & 14.7 & \textbf{57.3} & \textbf{78.7} & +48.7 & +64.0 \\
    OmniQuant    & 6.7  & 10.7 & \textbf{32.7} & \textbf{52.7} & +26.0 & +42.0
                 & 17.3 & 27.3 & \textbf{54.7} & \textbf{74.0} & +37.3 & +46.7 \\
    \bottomrule
  \end{tabular*}

  \vspace{0.8em}

   \caption{Planning success (\%) on LeWM TwoRoom after the first (iter 1) and second (iter 2) planning
           iteration, mean over 3 seeds. $\Delta$ is +WM-Quant minus the base method.}
  \label{tab:tworoom_mean}
  \vspace{-3mm}
  \begin{tabular*}{\textwidth}{@{\extracolsep{\fill}} l cccccc cccccc}
    \toprule
    & \multicolumn{6}{c}{W3A3} & \multicolumn{6}{c}{W4A4} \\
    \cmidrule(lr){2-7} \cmidrule(lr){8-13}
    & \multicolumn{2}{c}{Base} & \multicolumn{2}{c}{+WM-Quant} & \multicolumn{2}{c}{$\Delta$}
      & \multicolumn{2}{c}{Base} & \multicolumn{2}{c}{+WM-Quant} & \multicolumn{2}{c}{$\Delta$} \\
    \cmidrule(lr){2-3} \cmidrule(lr){4-5} \cmidrule(lr){6-7}
    \cmidrule(lr){8-9} \cmidrule(lr){10-11} \cmidrule(lr){12-13}
    Method & iter 1 & iter 2 & iter 1 & iter 2 & iter 1 & iter 2 & iter 1 & iter 2 & iter 1 & iter 2 & iter 1 & iter 2 \\
    \midrule
    FP           & 84.0 & 86.0 & -- & -- & -- & -- & 84.0 & 86.0 & -- & -- & -- & -- \\
    \midrule
    SmoothQuant  & 20.7 & 24.7 & \textbf{24.0} & \textbf{30.7} & +3.3  & +6.0
                 & 28.7 & 35.3 & \textbf{70.0} & \textbf{84.0} & +41.3 & +48.7 \\
    DuQuant      & 18.0 & 22.0 & \textbf{32.0} & \textbf{42.0} & +14.0 & +20.0
                 & 42.7 & 54.7 & \textbf{55.3} & \textbf{72.7} & +12.7 & +18.0 \\
    DuQuant+GPTQ & 18.7 & 26.0 & \textbf{38.7} & \textbf{51.3} & +20.0 & +25.3
                 & 40.0 & 50.7 & \textbf{57.3} & \textbf{72.0} & +17.3 & +21.3 \\
    RTN          & 16.7 & 24.0 & \textbf{24.0} & \textbf{33.3} & +7.3  & +9.3
                 & 28.7 & 37.3 & \textbf{70.7} & \textbf{84.0} & +42.0 & +46.7 \\
    OmniQuant    & \textbf{21.3} & 24.7 & 20.0 & \textbf{27.3} & $-$1.3 & +2.7
                 & 28.0 & 32.7 & \textbf{38.0} & \textbf{48.0} & +10.0 & +15.3 \\
    \bottomrule
  \end{tabular*}
\end{table*}

The remaining error after removing $\zeta_t$ and mitigating $\eta_t$ originates from the predictor. At each step, $
\epsilon_t$ is introduced, and then propagated throughout the remainder of the rollout. Standard static PTQ calibration minimizes the error for a single predictor step, but does not account for the propagation shown in Eq.~\eqref{eq:unrolled}. This means an optimal per-step approach may not necessarily be optimal for reducing error in the final rollout prediction. Moreover, typical PTQ methods calibrate the ranges of the encoder and predictor separately, with both aiming to minimize the distance between the quantized and non-quantized tensor output. This means the predictor’s ranges are fit to the activation distribution produced from the full-precision encoded state, without accounting for the initial encoder error. To address these issues, WM-Quant calibrates the model with respect to the entire rollout, essentially treating a rollout as a call to one deep model with an encoder feeding into a number of predictor blocks that share weights.

\begin{figure}[t]
  \centering
  \includegraphics[width=\linewidth, clip]{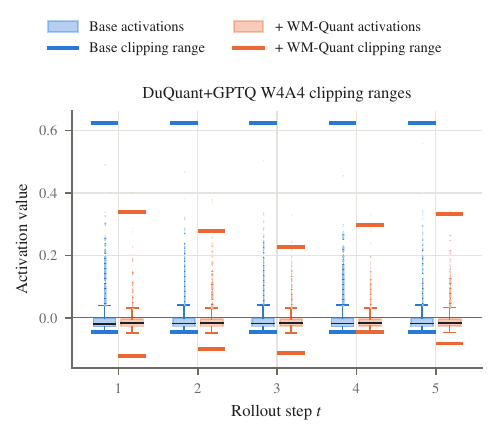}
  
  \caption{Activation distributions and clipping ranges across rollout steps (LeWM, PushT, W4A4) entering one quantized MLP predictor layer. Bars show the clipping range used by the activation quantizer: min/max calibration for the base method, and the range learned by rollout-aware scaling for WM-Quant.}
  \label{fig:ranges}
\end{figure}

Additionally, standard PTQ assigns each activation in the predictor a single set of per-tensor ranges that is shared throughout each rollout iteration. This can introduce error due to varying activation ranges at different rollout steps (Fig. \ref{fig:ranges}). WM-Quant assigns separate activation ranges to each rollout step, reducing the rate of error propagation. Because smoothing transformations and rotations are still shared, this change only introduces two scalar values per rollout step for each quantized unit (in our case, tensors).

Our calibration approach can be described as follows. Let $\Theta$ represent quantization ranges (encoder and predictor) and $\mathbb{E}_{\mathcal D}$ the average over a set of calibration rollouts recorded from the planner. WM-Quant calibrates with the following loss \ensuremath{\mathcal{L}(\Theta)} over an $n$-step rollout:
\begin{equation}
  \mathcal{L}(\Theta) = \mathbb{E}_{\mathcal D}\sum_{t=1}^{n} w_t\,\big\|\delta_{\tau+t}(\Theta)\big\|_2^2,
  \label{eq:loopq}
\end{equation}
where $w_t = t/\sum_{t'=1}^{n} t'$ weights each rollout step $t$, placing greater emphasis on later steps where error has accumulated. Both the encoder ranges and the per-step predictor rollout ranges are updated jointly with the Adam optimizer \cite{kingma2017adammethodstochasticoptimization} through the full quantized rollout, using the straight-through estimator \cite{bengio2013estimatingpropagatinggradientsstochastic} for rounding.

\section{Experiments}

\begin{table*}[!t]
  \centering

\caption{Planning success (\%) on DINO-WM Wall at each MPC iteration.  $\Delta$ is +WM-Quant minus the base method.}
  \label{tab:dinowm_wall}
  {\footnotesize
  \begin{tabular*}{\textwidth}{@{\extracolsep{\fill}} ll ccccc ccccc}
    \toprule
    & & \multicolumn{5}{c}{W3A3} & \multicolumn{5}{c}{W4A4} \\
    \cmidrule(lr){3-7} \cmidrule(lr){8-12}
    \multicolumn{2}{l}{MPC iteration} & 1 & 5 & 10 & 15 & 20 & 1 & 5 & 10 & 15 & 20 \\
    \midrule
    FP & & 48 & 94 & 94 & 94 & 94 & 48 & 94 & 94 & 94 & 94 \\
    \midrule
    \multirow{3}{*}{SmoothQuant}
      & Base       & 0 & 0  & 2  & \textbf{4}    & \textbf{4}    & 0 & \textbf{4}    & \textbf{4}    & 4 & 4 \\
      & +WM-Quant  & 0 & \textbf{2}  & 2  & 2    & 2    & 0 & 2    & 2    & 4 & 4 \\
      & $\Delta$   & 0 & +2 & 0  & $-$2 & $-$2 & 0 & $-$2 & $-$2 & 0 & 0 \\
    \midrule
    \multirow{3}{*}{DuQuant}
      & Base       & 0 & 6 & 6  & 6  & 6  & 0   & 8   & 16  & 20  & 24  \\
      & +WM-Quant  & 0 & 6 & \textbf{12} & \textbf{14} & \textbf{18}
                   & \textbf{22} & \textbf{88} & \textbf{92} & \textbf{92} & \textbf{92} \\
      & $\Delta$   & 0 & 0 & +6 & +8 & +12
                   & +22 & +80 & +76 & +72 & +68 \\
    \midrule
    \multirow{3}{*}{DuQuant+GPTQ}
      & Base       & 0   & 6   & 12  & 12  & 12  & 2   & 16  & 32  & 38  & 46  \\
      & +WM-Quant  & \textbf{10} & \textbf{24} & \textbf{38} & \textbf{44} & \textbf{44}
                   & \textbf{32} & \textbf{96} & \textbf{98} & \textbf{98} & \textbf{98} \\
      & $\Delta$   & +10 & +18 & +26 & +32 & +32
                   & +30 & +80 & +66 & +60 & +52 \\
    \midrule
    \multirow{3}{*}{RTN}
      & Base       & 0 & 0  & 0  & 0  & 0  & 0  & 0  & 0  & 2  & 4 \\
      & +WM-Quant  & 0 & \textbf{2}  & \textbf{2}  & \textbf{2}  & \textbf{2}
                   & \textbf{2}  & \textbf{4}  & \textbf{4}  & \textbf{4}  & 4 \\
      & $\Delta$   & 0 & +2 & +2 & +2 & +2
                   & +2 & +4 & +4 & +2 & 0 \\
    \midrule
    \multirow{3}{*}{OmniQuant}
      & Base       & 0 & 0 & 0 & 0 & 0 & 0 & 0 & \textbf{2}     & \textbf{2}     & \textbf{2}    \\
      & +WM-Quant  & 0 & 0 & 0 & 0 & 0 & 0 & 0 & 0     & 0     & 0     \\
      & $\Delta$   & 0 & 0 & 0 & 0 & 0 & 0 & 0 & $-$2  & $-$2  & $-$2  \\
    \bottomrule
  \end{tabular*}
  }

  \vspace{0.8em}

\begin{minipage}[t]{0.47\textwidth}
    \centering
    \caption{Planning success (\%) on DINO-WM PushT at W4A4 at each MPC iteration. $\Delta$ is
             +WM-Quant minus the base method.}
    \label{tab:dinowm_pusht_w4a4}
    \footnotesize
    \begin{tabular*}{\linewidth}{@{\extracolsep{\fill}} ll ccc}
      \toprule
      \multicolumn{2}{l}{MPC iteration} & 1 & 5 & 10 \\
      \midrule
      FP & & 84 & 90 & 90 \\
      \midrule
      \multirow{3}{*}{SmoothQuant}
        & Base      & 0 & 0 & 0 \\
        & +WM-Quant & 0 & 0 & 0 \\
        & $\Delta$  & 0 & 0 & 0 \\
      \midrule
      \multirow{3}{*}{DuQuant}
        & Base      & 0  & 0  & 0   \\
        & +WM-Quant & \textbf{6}  & \textbf{8}  & \textbf{12}  \\
        & $\Delta$  & +6 & +8 & +12 \\
      \midrule
      \multirow{3}{*}{DuQuant+GPTQ}
        & Base      & 0 & 2 & 2 \\
        & +WM-Quant & \textbf{28} & \textbf{36} & \textbf{42} \\
        & $\Delta$  & +28 & +34 & +40 \\
      \midrule
      \multirow{3}{*}{RTN}
        & Base      & 0 & 0 & 0 \\
        & +WM-Quant & 0 & \textbf{2} & \textbf{2} \\
        & $\Delta$  & 0 & +2 & +2 \\
      \midrule
      \multirow{3}{*}{OmniQuant}
        & Base      & 0 & 0 & 0 \\
        & +WM-Quant & 0 & 0 & 0 \\
        & $\Delta$  & 0 & 0 & 0 \\
      \bottomrule
    \end{tabular*}
  \end{minipage}%
  \hfill
      \begin{minipage}[t]{0.50\textwidth}
    \centering
    \footnotesize
    \setlength{\tabcolsep}{4pt}
    \caption{Ablation of WM-Quant on LeWM PushT, DuQuant+GPTQ at W3A3. Planning success (\%)
             after two MPC iterations, mean $\pm$ standard error over 10 seeds.
             $\Delta$: change relative to full WM-Quant, with a 95\% paired $t$-interval over seeds. Random 3 encoder blocks and random preserved layers give the same or higher bit-budgets as early-encoder precision and control-path preservation respectively.}
    \label{tab:ablation}
    \begin{tabular}{l ccc}
      \toprule
      Method & Success & $\Delta$ & 95\% CI \\
      \midrule
      WM-Quant (full)                         & \textbf{56.0} \std{0.9} & --      & -- \\
      \midrule
      \multicolumn{4}{l}{\textit{Removing a component}} \\
      \quad w/o control-path preservation     & 24.4 \std{1.6} & \boldmath$-$31.6 & [$-$35.3, $-$27.9] \\
      \quad w/o early-encoder precision       & 52.8 \std{2.3} & \boldmath$-$3.2  & [$-$7.6, 1.2] \\
      \quad w/o rollout-aware scaling         & 36.4 \std{1.2} & \boldmath$-$19.6 & [$-$23.2, $-$16.0] \\
      \midrule
      \multicolumn{4}{l}{\textit{Mixed-precision tests}} \\
      \quad random preserved layers           & 27.6 \std{2.0} & \boldmath$-$28.4 & [$-$33.0, $-$23.8] \\
      \quad random 3 encoder blocks              & 55.6 \std{1.9} & \boldmath$-$0.4  & [$-$4.4, 3.6] \\
      \midrule
      \multicolumn{4}{l}{\textit{Rollout-aware scaling variants}} \\
      \quad no training (initial ranges)      & 34.6 \std{2.0} & \boldmath$-$21.4 & [$-$25.2, $-$17.6] \\
      \quad no encoder-range training         & 50.2 \std{2.2} & \boldmath$-$5.8  & [$-$10.7, $-$0.9] \\
      \quad ranges shared across steps        & 49.2 \std{4.0} & \boldmath$-$6.8  & [$-$15.3, 1.7] \\
      \midrule
      DuQuant+GPTQ (no WM-Quant)              & 9.6 \std{1.1}  & \boldmath$-$46.4 & [$-$50.0, $-$42.8] \\
      \bottomrule
    \end{tabular}
  \end{minipage}
\end{table*}

\begin{table*}[!htb]
\centering
\caption{W4A4 integer kernels on a Jetson AGX Orin, each method with and without WM-Quant (WMQ); FP16 is
run with the default PyTorch implementation. Planning: one CEM cost evaluation of 300 candidate action sequences, end-to-end (encoding the
start and goal observations, the 5-step rollout, the cost). Speedup is relative to FP16. Predictor: the 5 predictor calls alone (LeWM: with
their heads). Encoder: 50 images. Peak memory measures the additional memory allocated, on top of weights and inputs, during one full planning call.}
\label{tab:kernels}
\vspace{-2mm}
\setlength{\tabcolsep}{4pt}
\resizebox{\textwidth}{!}{%
\begin{tabular}{cl rr rr rr rr rr rr rr rr}
\toprule
 & & \multicolumn{2}{c}{\shortstack{Planning\\(ms)}} & \multicolumn{2}{c}{\shortstack{Speedup\\($\times$)}} & \multicolumn{2}{c}{\shortstack{Predictor\\(ms)}} & \multicolumn{2}{c}{\shortstack{Encoder\\(ms)}} & \multicolumn{2}{c}{\shortstack{Pred.\ weights\\(MiB)}} & \multicolumn{2}{c}{\shortstack{Enc.\ weights\\(MiB)}} & \multicolumn{2}{c}{\shortstack{Peak memory\\(MiB)}} & \multicolumn{2}{c}{\shortstack{Enc.\ activations\\(MiB)}} \\
\cmidrule(lr){3-4}\cmidrule(lr){5-6}\cmidrule(lr){7-8}\cmidrule(lr){9-10}\cmidrule(lr){11-12}\cmidrule(lr){13-14}\cmidrule(lr){15-16}\cmidrule(lr){17-18}
 & Method & base & +WMQ & base & +WMQ & base & +WMQ & base & +WMQ & base & +WMQ & base & +WMQ & base & +WMQ & base & +WMQ \\
\midrule
\multirow{6}{*}{\rotatebox[origin=c]{90}{DINO-WM}} & FP16 & \multicolumn{2}{c}{3877} & \multicolumn{2}{c}{1.00} & \multicolumn{2}{c}{3756} & \multicolumn{2}{c}{37.9} & \multicolumn{2}{c}{38.7} & \multicolumn{2}{c}{43.5} & \multicolumn{2}{c}{3847} & \multicolumn{2}{c}{101.4} \\
 & RTN & 1824 & 1854 & 2.12 & 2.09 & 1761 & 1795 & 31.7 & 33.1 & 12.7 & 12.7 & 13.6 & 16.1 & 1649 & 1649 & 44.8 & 46.9 \\
 & SmoothQuant & 1822 & 1853 & 2.13 & 2.09 & 1763 & 1794 & 31.5 & 32.8 & 12.7 & 12.7 & 13.6 & 16.1 & 1649 & 1649 & 44.8 & 46.9 \\
 & OmniQuant & 1825 & 1851 & 2.12 & 2.09 & 1763 & 1794 & 31.7 & 33.1 & 12.8 & 12.8 & 13.6 & 16.1 & 1649 & 1649 & 44.8 & 47.0 \\
 & DuQuant & 2525 & 2410 & 1.54 & 1.61 & 2454 & 2343 & 43.7 & 44.8 & 14.5 & 14.5 & 17.0 & 19.5 & 2682 & 2683 & 69.0 & 84.4 \\
 & DuQuant+GPTQ & 2523 & 2408 & 1.54 & 1.61 & 2458 & 2344 & 43.5 & 44.7 & 14.5 & 14.5 & 17.0 & 19.5 & 2683 & 2683 & 67.1 & 83.4 \\
\midrule
\multirow{6}{*}{\rotatebox[origin=c]{90}{LeWM}} & FP16 & \multicolumn{2}{c}{78.6} & \multicolumn{2}{c}{1.00} & \multicolumn{2}{c}{52.3} & \multicolumn{2}{c}{23.3} & \multicolumn{2}{c}{22.09} & \multicolumn{2}{c}{12.00} & \multicolumn{2}{c}{31.6} & \multicolumn{2}{c}{61.5} \\
 & RTN & 48.2 & 49.6 & 1.63 & 1.58 & 34.3 & 34.2 & 20.2 & 20.6 & 7.15 & 8.77 & 4.45 & 4.94 & 32.3 & 41.6 & 48.2 & 48.2 \\
 & SmoothQuant & 48.0 & 49.5 & 1.64 & 1.59 & 34.3 & 34.5 & 20.3 & 20.6 & 7.15 & 8.77 & 4.45 & 4.94 & 32.6 & 41.7 & 48.2 & 48.3 \\
 & OmniQuant & 48.0 & 49.8 & 1.64 & 1.58 & 34.3 & 34.4 & 20.3 & 20.6 & 7.22 & 8.84 & 4.45 & 4.94 & 32.6 & 41.7 & 48.2 & 48.2 \\
 & DuQuant & 56.5 & 58.0 & 1.39 & 1.36 & 40.0 & 40.2 & 32.7 & 32.7 & 9.06 & 10.68 & 7.85 & 8.62 & 37.8 & 41.7 & 48.2 & 58.8 \\
 & DuQuant+GPTQ & 56.5 & 57.7 & 1.39 & 1.36 & 39.9 & 40.2 & 32.7 & 32.7 & 9.06 & 10.68 & 7.85 & 8.62 & 38.2 & 41.7 & 48.2 & 57.7 \\
\bottomrule
\end{tabular}}
\end{table*}

\begin{table}[t]
  \centering
  \caption{Early-encoder precision on DINO-WM Wall at W3A3 (DuQuant+GPTQ + WM-Quant): planning success (\%)
           at each MPC iteration, averaged over 3 seeds. Rows give which encoder blocks keep 8-bit activations;
           \emph{Random 3} gives the same budget to three randomly chosen blocks. $\Delta$ is the variant minus
           \emph{First 3}.}
  \label{tab:dinowm_wall_early}
  \footnotesize
  \setlength{\tabcolsep}{4pt}
  \begin{tabular*}{\linewidth}{@{\extracolsep{\fill}} l cccc}
    \toprule
    MPC iteration & 1 & 5 & 10 & 20 \\
    \midrule
    FP & 48 & 94 & 94 & 94 \\
    \midrule
    First 3 (WM-Quant) & 2.7 & \textbf{15.3} & \textbf{26.7} & \textbf{36.7} \\
    Random 3           & \textbf{3.3} & 8.7 & 18.7 & 26.7 \\
    None               & 0.0 & 6.0 & 8.0 & 14.0 \\
    \midrule
    $\Delta$ Random 3  & +0.7   & $-$6.7 & $-$8.0  & $-$10.0 \\
    $\Delta$ None      & $-$2.7 & $-$9.3 & $-$18.7 & $-$22.7 \\
    \bottomrule
  \end{tabular*}
\end{table}

\subsection{Experimental settings}
\label{sec:exp_settings}
We evaluate WM-Quant on two state-of-the-art JEPA-style world models, LeWM \cite{maes2026leworldmodelstableendtoendjointembedding} and DINO-WM \cite{zhou2025dinowm}. Both encode image observations into latent states, but their encoders differ substantially. LeWM uses a small ViT \cite{dosovitskiy2021an} trained from scratch jointly with the predictor and represents each frame by a single vector, its CLS token. DINO-WM instead uses a frozen, pretrained DINOv2 \cite{oquab2024dinov2learningrobustvisual} encoder and represents each frame as a grid of patch tokens (196 in our experiments). The two models also condition the predictor on control inputs differently: DINO-WM concatenates action and proprioceptive features to every patch token, whereas LeWM injects actions through AdaLN \cite{Peebles_2023_ICCV} modulation. We evaluate LeWM on PushT \cite{florence2021implicit, 10.1177/02783649241273668}, a contact-based manipulation task, and on TwoRoom \cite{sobal2025learning}, a simpler navigation task. We evaluate DINO-WM on PushT and Wall \cite{zhou2025dinowm}, a navigation task with a wall obstacle. We test on W3A3 and W4A4 because extremely low bit-widths lead to severe degradation and more measurable error propagation. Only one seed was used for DINO-WM main results due to the increased number of required MPC iterations and compute time compared to LeWM.

\subsection{WM-Quant results}
\paragraph{LeWM results}
As shown in Table~\ref{tab:pusht_mean} for PushT and Table~\ref{tab:tworoom_mean} for TwoRoom,
WM-Quant substantially improves LeWM planning success across all PTQ base methods.
Per-seed results are provided in the supplementary material. WM-Quant is even able to recover a significant average increase of 57.34 planning success percentage points on LeWM PushT W4A4 across PTQ methods, achieving near floating-point results with little overhead. While more PTQ method-dependent, WM-Quant is also able to recover near floating-point results on LeWM TwoRoom W4A4. 
Even on the W3A3 setting, where weights and activations are limited to eight representable values, WM-Quant delivers success rates up to 58\% on PushT and 51.3\% on TwoRoom. These results indicate that mitigating error propagation as discussed in Sections \ref{sec:error_prop} and \ref{sec:WM-Quant} substantially increases downstream planning success.

\paragraph{DINO-WM results}
As displayed in Table~\ref{tab:dinowm_wall} for Wall and Table~\ref{tab:dinowm_pusht_w4a4} for PushT, WM-Quant is able to recover a significant portion of planning success on both datasets when rotational transformations are applied across the weights and activations via DuQuant. The disparity across quantization methods can likely be attributed to extreme outliers in the activations of the pretrained frozen DINOv2 encoder \cite{oquab2024dinov2learningrobustvisual}, which drastically increase quantization sensitivity at lower bit-widths \cite{kim2026activationquantizationvisionencoders}. This makes it substantially more difficult for WM-Quant to recover accuracy alone, without rotational transformations to smooth extreme outliers across channels. With these transformations applied, WM-Quant is able to recover near floating-point accuracy on Wall W4A4 and substantially improve results on PushT W4A4. WM-Quant also increases planning success for W3A3 by an average of 22 percentage points on Wall with DuQuant-based methods. For RTN, SmoothQuant, and OmniQuant, the base quantized DINO-WM models achieve near-zero success, suggesting that their errors are too large for planning to succeed within the episode horizon even with WM-Quant.

\paragraph{Empirical validation of WM-Quant}
Table \ref{tab:ablation} ablates each core component of WM-Quant on LeWM PushT with DuQuant+GPTQ as the base PTQ method at W3A3. WM-Quant increases planning success by 46.4 percentage points. Component-wise analysis of WM-Quant reveals that rollout-aware scaling and control-path preservation are both essential to planning success. Early-encoder precision and per-rollout step scales also improve the accuracy and the stability of the WM-Quant framework at little cost, but admittedly provide more modest gains on LeWM. However, as seen in DINO-WM W3A3 Wall encoder ablations in Table \ref{tab:dinowm_wall_early}, early-encoder precision provides a much greater leap in accuracy for DINO-WM, due to its hard-to-quantize pretrained DINOv2 encoder.

\paragraph{Deployment efficiency}
To validate our claims of limited additional latency or memory overhead from WM-Quant, we developed custom CUDA kernels on LeWM and DINO-WM on all of the evaluated quantization methods. We report latency and memory usage measured on the NVIDIA Jetson AGX Orin in {Table~\ref{tab:kernels}}. We leave the selection of ideal bit-widths for particular edge devices to future work. Over one CEM iteration, across both models and all quantization methods, we see an average speedup of 1.71$\times$ without WM-Quant, which only decreases to 1.70$\times$ with WM-Quant. The control-path preservation in the predictor leads to no increase in predictor weight or activation memory usage for all DINO-WM scenarios, due to padding in the GEMM kernel and a small separate buffer for the control channels. In LeWM it leads to small increases in both. Both models see an increase in the memory usage of encoder weights and activations, though it is nowhere near rivaling the peak memory usage of DINO-WM, and it has a lower maximum memory usage during planning than the LeWM predictor (Table~\ref{tab:kernels} tests 50 images for the Enc. activations column) meaning this increase does not dramatically affect deployability. Notably, WM-Quant decreases planning latency in DINO-WM planning when DuQuant is used, due to the removal of control channels reducing the required number of DuQuant rotation blocks.

\section{Conclusion}
In this work, we present WM-Quant, to our knowledge, the first quantization framework for JEPA-style planning-based world models. We conduct a detailed analysis of how quantization error propagates throughout world models and introduce low-cost methods that directly target each error type. By keeping early encoder layers and control information at higher precisions, along with calibrating per-loop scales and encoder scales over the entire rollout, we are able to create a quantization scheme that reduces rollout error dramatically. On LeWM we demonstrate an average success rate improvement of 39.4 percentage points on W4A4 and 21.8 percentage points on W3A3 across datasets. On DINO-WM we see improvements of up to 80 percentage points on some iterations and configurations. Additionally, we perform comprehensive ablations to empirically justify our approach.


\section{Acknowledgments}
This work was supported in part by the National Science Foundation (NSF) under Grant No. 2340249. The paper also includes contributions from an undergraduate researcher supported by the NSF Research Experiences for Undergraduates (REU) program under Grant No. CCF-2621730. The authors gratefully acknowledge the computational resources provided by the Theia high performance computing cluster at the University of South Carolina which is supported by National Science Foundation Major Research Instrumentation Grant No. 2320292. We also acknowledge the technical assistance and resources provided by Research Computing at the University of South Carolina (RRID:SCR\_0274388).
{
    \small
    \bibliographystyle{ieeenat_fullname}
    \bibliography{main}
}


\end{document}